\documentclass[sigconf]{acmart}

\usepackage{booktabs} % For formal tables

\usepackage{amssymb}
\usepackage{soul}
\usepackage{bm}

\usepackage{multirow}
\usepackage{rotating}

\usepackage{tikz}
\usetikzlibrary{arrows,snakes,backgrounds}
 
\usepackage{algorithm}
\usepackage{algorithmicx}
\usepackage[noend]{algpseudocode}

\algrenewcommand\algorithmiccomment[2][\itshape]{{#1\hfill\(\triangleright\)
    #2}}
\algrenewcommand{\algorithmicrequire}{\textbf{Input:}}
\algrenewcommand{\algorithmicensure}{\textbf{Output:}}

\usepackage{amsmath}
\usepackage{todonotes}

\newcommand{\cX}{{\mathcal{X}}}
\newcommand{\cY}{{\mathcal{Y}}}

\newcommand{\cG}{{\mathcal{G}}}
\newcommand{\argmin}{\operatornamewithlimits{argmin}}
 \newcommand{\given}{\, | \,}
  \newcommand{\with}{\, | \,}
\newcommand{\set}[1]{\mathcal{#1}}
\renewcommand{\vec}[1]{#1}

\setcopyright{rightsretained}
\acmYear{2017}
\copyrightyear{2017}
\begin{document}
\title{Predicting Rankings of Software Verification Competitions}
\titlenote{This work was partially supported by the German Research Foundation (DFG) within the Collaborative Research Centre ``On-The-Fly Computing'' (SFB 901).}
%\subtitle{Extended Abstract}
%\subtitlenote{The full version of the author's guide is available as \texttt{acmart.pdf} document}

%\authornote{The secretary disavows any knowledge of this author's actions.}
%\affiliation{%
  %\institution{Institute for Clarity in Documentation}
  %\streetaddress{P.O. Box 1212}
%\email{webmaster@marysville-ohio.com}

\author{Mike Czech, Eyke H{\"{u}}llermeier, Marie-Christine Jakobs, Heike Wehrheim  }
\affiliation{\institution{Department of Computer Science\\
Paderborn University \\ Germany}}
%\email{jsmith@affiliation.org}

%\author{Eyke H{\"{u}}llermeier}
%\affiliation{\institution{Department of Computer Science\\
%Paderborn University \\ Germany}}
%%\email{jpkumquat@consortium.net}
%
%\author{Heike Wehrheim}
%\affiliation{\institution{Department of Computer Science\\
%Paderborn University \\ Germany}}
%\email{jpkumquat@consortium.net}


\begin{abstract}
Software verification competitions, such as the annual SV-COMP, evaluate software verification tools  with respect to their effectivity and efficiency. Typically, the outcome of a competition is  a (possibly category-specific) {\em ranking} of the tools. For many applications, such as building portfolio solvers, it would be desirable to have an idea of the (relative) performance of verification tools on a given verification task \emph{beforehand}, i.e., prior to actually running all tools on the task.

In this paper, we present a machine learning approach to {\em predicting} rankings of tools on verification tasks. The method builds upon so-called label ranking algorithms, which we complement with appropriate {\em kernels} providing a similarity measure for verification tasks. Our kernels employ a graph representation for software source code that mixes elements of control flow and program dependence graphs with abstract syntax trees. Using data sets from SV-COMP, we demonstrate our rank prediction technique to generalize well and achieve a rather high predictive accuracy. In particular, our method outperforms a recently proposed feature-based approach of Demyanova et al.\ (when applied to rank predictions).
\end{abstract}

 \begin{CCSXML}
<ccs2012>
<concept>
<concept_id>10010147.10010257.10010258.10010259.10003268</concept_id>
<concept_desc>Computing methodologies~Ranking</concept_desc>
<concept_significance>500</concept_significance>
</concept>
<concept>
<concept_id>10010147.10010257.10010293.10010075.10010295</concept_id>
<concept_desc>Computing methodologies~Support vector machines</concept_desc>
<concept_significance>300</concept_significance>
</concept>
<concept>
<concept_id>10010147.10010257.10010339</concept_id>
<concept_desc>Computing methodologies~Cross-validation</concept_desc>
<concept_significance>100</concept_significance>
</concept>
<concept>
<concept_id>10011007.10010940.10010992.10010998.10010999</concept_id>
<concept_desc>Software and its engineering~Software verification</concept_desc>
<concept_significance>100</concept_significance>
</concept>
<concept>
<concept_id>10011007.10011074.10011099.10011692</concept_id>
<concept_desc>Software and its engineering~Formal software verification</concept_desc>
<concept_significance>100</concept_significance>
</concept>
</ccs2012>
\end{CCSXML}

\ccsdesc[500]{Computing methodologies~Ranking}
\ccsdesc[300]{Computing methodologies~Support vector machines}
\ccsdesc[100]{Computing methodologies~Cross-validation}
\ccsdesc[100]{Software and its engineering~Software verification}
\ccsdesc[100]{Software and its engineering~Formal software verification}

\keywords{Software verification, machine learning, ranking. }

\maketitle

\section{Introduction}

The annual holding of software verification competitions has recently stimulated the development of verification tools, in particular the tuning of tools towards performance and precision. The participating candidate tools typically employ a large range of different techniques, from static analysis, abstract interpretation and automata-based techniques to SAT or SMT solving.  In the area of automatic verification, the most prominent competition today is the Competition on Software Verification SV-COMP \cite{DBLP:conf/tacas/000115}. Over the years, the SV-COMP community has collected a large number of benchmark verification tasks, i.e., software source code together with properties to be proven, and is constantly continuing to do so. Verification tasks are bundled in categories, and the outcome of SV-COMP are {\em rankings} (overall and per category) computed by means of a scoring schema. 

Rankings of verification tools on verification tasks in software competitions offer an {\em a-posteriori} insight into the particular usefulness of a tool on a verification task. However, for programmers wanting to select a tool for verification of their program, or for building {\em portfolio solvers}, a prediction on  a likely ranking {\em without} actually running (all or some) tools is needed. In this paper, we propose such a method.   

Our method for rank prediction builds upon machine learning techniques, more precisely on so-called \emph{label ranking} algorithms \cite{DBLP:books/daglib/p/FurnkranzH10a} using support vector machines \cite{DBLP:conf/colt/BoserGV92} as base learners. As training data, we take the SV-COMP results of 2015 and the only recently published results of 2017.  The key ingredient of our approach is the definition of a number of {\em kernels} \cite{DBLP:books/daglib/0026002}, which act as similarity measures on verification tasks. So far, two other machine learning  methods for selecting tools or algorithms for verification have been proposed \cite{DBLP:conf/msr/TulsianKKLN14,DBLP:conf/cav/DemyanovaPVZ15}, both of them being based on feature vectors: while Tulsian et al.~\cite{DBLP:conf/msr/TulsianKKLN14} only employ structural features of programs (like the number of arrays, loops, recursive functions), Demyanova et al.~\cite{DBLP:conf/cav/DemyanovaPVZ15} uses a number of data-flow analyses to also determine more sophisticated features (e.g., certain loop patterns).  Thus, both approaches try to explicitly capture aspects  of source code that make verification hard (for some or all tools). With our kernels, we take a different approach: we supply the learning algorithm with a representation of source code that enables the learner itself to identify the distinguishing patterns. We believe that our kernels are thus more readily usable for other program analysis tasks, for which a machine learning method might be considered (e.g., program classification or program analysis). In  that case, we could use exactly the same kernel and just exchange  the training data. Still, our experiments show that the prediction accuracy for rankings of software verification competitions is higher than that of Demyanova et al.~\cite{DBLP:conf/cav/DemyanovaPVZ15} (when using their feature vectors for predicting rankings, not just predicting winners)\footnote{A comparison with Tulsian et al.~\cite{DBLP:conf/msr/TulsianKKLN14} is difficult due to non-reproducability of their results.}. 

More specifically, our kernels are constructed on {\em graph representations} of source code. We have experimented with different (weighted) combinations of control flow graphs (CFGs), program dependence graphs (PDGs), and abstract syntax trees (ASTs). In these, concrete inscriptions on nodes (like \verb#x:=y+1#) are first of all replaced by abstract {\em labels} (e.g., \verb+Assign+). Such labelled graphs are used within our specific adaptation of the Weisfeiler-Lehman graph kernel framework \cite{DBLP:journals/jmlr/ShervashidzeSLMB11} that compares graphs not only according to their labels (and how often they occur) but also according to {\em associations}  between labels (via edges in the graph).  This is achieved by iteratively comparing  larger and larger subtrees of nodes, where the maximum  depth of subtrees to be considered is a parameter to the framework. The choice of Weisfeiler-Lehman kernels is motivated by their better scalability compared to other graph kernels, such as random walk or shortest path kernels (see \cite{DBLP:journals/jmlr/ShervashidzeSLMB11}). We prove our kernels to be positive definite, which is the key property for kernels to be usable for machine learning.  
The ranking is finally computed by a method for rank prediction via pairwise comparison \cite{DBLP:books/daglib/p/FurnkranzH10a}, using support vector machines as base learners. 

We have implemented our technique and carried out experimental (cross-validation) studies using data from SV-COMP 2015 and 2017. The experiments show that our technique can predict rankings with a rather high accuracy, using Spearman's rank correlation \cite{Spear} to compare predicted with true rankings. To see how our technique compares to existing approaches, we have also used the feature vectors of Demyanova et al.~\cite{DBLP:conf/cav/DemyanovaPVZ15} for rank prediction. It turns out that, for three different data sets (containing general safety, termination and memory safety verification tasks), our technique outperforms the technique of \cite{DBLP:conf/cav/DemyanovaPVZ15} in almost all instantiations (choosing a depth for subtrees and a combination of CFG, PDG and AST) .

\smallskip
\noindent 
Summarizing, this paper makes the following contributions:
\begin{itemize}
\item We propose a technique for the prediction of {\em rankings} in software verification competitions (instead of just predicting winners);
\item we present an expressive representation of source code ready for use in other machine learning approaches to program analysis;  
\item we experimentally demonstrate our technique---despite  being more general and more widely applicable---to compare favorably with existing approaches on the specific task of predicting rankings of software verification tools.

\end{itemize}

\noindent All data of 2015 and software is available at \url{https://github.com/zenscr/PyPRSVT}.

\section{Representing Verification Tasks}

Our objective is to {\em predict} rankings of software verification competitions via machine learning. To this end, the learning algorithm has to be supplied with training data, which, in our case, is readily available from the SV-COMP website. We start with explaining what kind of data our rank prediction technique is supplied with, and how this data is represented.

The purpose of SV-COMP is to compare verification tools with respect to their effectiveness and efficiency. To this end, the tools are supplied with verification tasks.

\begin{figure}[t]
\begin{verbatim}
1  int i;         6  i = 0; 						
2  int n;	        7  while (i <= n) 					
3  int sn;        8       sn = sn + 2; 					
4  n = input();   9       i = i + 1;  				
5  sn = 0;        10 assert (sn == n*2 || sn == 0); 
\end{verbatim} 
\caption{The verification task $P_{\mathit{SUM}}$}
\label{fig:sum}
\end{figure}

\begin{definition} A {\em verification task} $(P,\varphi)$ consists of a program $P$ (for SV-COMP written in C) and a property (also called  specification) $\varphi$ (typically written as assertion into the program).
\end{definition}

\begin{table*} [t] 
\begin{tabular}{||c|l||c|l||}
\hline
\texttt{Loop} & loop & 
\texttt{If} & conditional \\
\hline
\texttt{Decl} & variable declaration & \texttt{Assign} & variable assignment \\
\hline 
\texttt{Incr} & variable increment  & \texttt{Assert} & assertion \\ 
\hline 
\texttt{Ref} & variable reference & \texttt{Int\_Literal\_Small} & integer literal in [0,10] \\
\hline
\texttt{Function\_Call} & function call & \texttt{Int\_Literal\_Medium} & integer literal in [10,100] \\
\hline
\texttt{Function\_Return} & function return & \texttt{Int\_Literal\_Large} & integer literal $>$ 100  \\
\hline 
\end{tabular}
\vspace*{.1cm}
\caption{Some node identifiers and their meaning}
\label{tab:ids}
\end{table*}

\noindent Figure \ref{fig:sum} shows our running example $P_{\mathit{SUM}}$ of a verification task (computing $n$ times 2 via addition). In a verification run, a verification tool is run on a verification task in order to determine whether the program fulfills the specification. The outcome of such a verification run is a pair (TIME, ANSWER),\footnote{In addition, witnesses are part of the outcomes. Witnesses have only been part of the scoring scheme of 2017, and are thus for reasons of harmonisation of 2015 and 2017 not considered here. } where TIME is the time in seconds from the start of the verification run to its end, and ANSWER is of the following form:
\begin{description}
\item[TRUE] when the verification tool has concluded that $P$ satisfies $\varphi$, 
\item[FALSE] when the verification tool has concluded that $P$ violates $\varphi$, and 
\item[UNKNOWN] when no conclusive result was achieved. 
\end{description}  

\noindent In SV-COMP, verification tasks are bundled into categories (e.g., memory safety, termination, concurrency). Ranking of tools is first of all being carried out per category (but then extended to meta categories). The ranking within a category is done via a scoring schema which gives positive and negative points to outcomes, e.g., negative points  when the tool incorrectly concluded the property to be valid for the program. When the scores of two tools are the same, the runtimes (of successful runs) determine the ordering. The data from SV-COMP available for learning rankings thus consists of verification tasks in different categories, outcomes of tools on these tasks and scores assigned to these outcomes as well as the final ranking.

The purpose of the machine learning algorithm is to learn from these observations how tools will perform on specific verification tasks. Our machine learning technique is based on {\em kernel methods} (see e.g.\ \cite{DBLP:books/daglib/0026002}). In general, a kernel can be interpreted as a similarity measure on data instances (in our case verification tasks), with the idea that similar results (in our case rankings) are produced for similar instances. While kernel-based learning algorithms are completely generic, the kernel function itself is application-specific and, to achieve strong performance, needs to be designed in an appropriate way. In other words, a key question is how to define kernels suitable for the application at hand. 

The simplest way of defining a kernel is via the inner product of {\em feature vectors}, i.e., vectorial representations of data objects. In the two approaches existing so far \cite{DBLP:conf/msr/TulsianKKLN14,DBLP:conf/cav/DemyanovaPVZ15}, corresponding features of programs, such as the number of loops, conditionals, pointer variables, or arrays in a program,  are defined in an explicit way. Obviously, this approach requires sufficient domain knowledge to identify those features that are important for the prediction problem at hand. Our approach essentially differs in that features are specified in a more indirect way, namely by systematically extracting (a typically large number of) generic features from a suitable representation of the verification task. Selecting the useful features and combining them appropriately is then basically left to the learner.

\begin{figure*}[t]
\begin{center}
\includegraphics[width=0.65\textwidth]{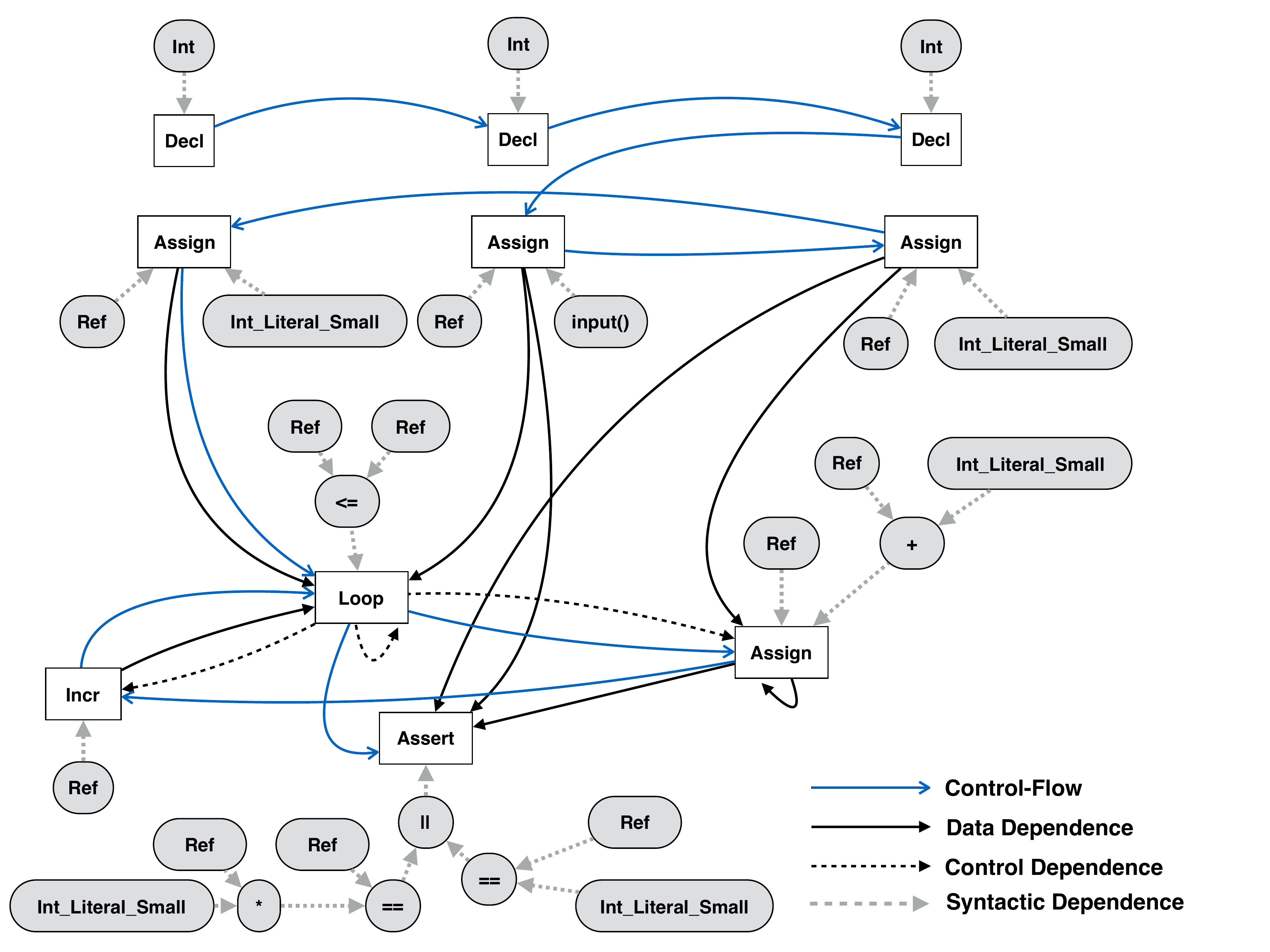}
\end{center}
\caption{Graph representation of $P_{\mathit{SUM}}$ eliding labelling $\nu$}
\label{fig:graph-sum}
\end{figure*}

But how to represent the verification tasks in a proper way? The first idea is to use the source code itself (i.e., strings). However, the source code of two programs might look very different although the underlying program is actually the same (different variable names, \texttt{while} instead of \texttt{for} loops, etc.). What we need is a representation that abstracts from issues like variable names but still represents the structure of programs, in particular dependencies between elements of the program. These considerations (and some experiments comparing different representations) have led to a {\em graph} representation of programs combining concepts of three existing program representations:
\begin{description}
\item[Control flow graphs:] CFGs record the control flow in programs and thus the overall structure with loops, conditionals etc.; these are needed, for example, to see loops in programs. 
\item[Program dependence graphs:] PDGs \cite{DBLP:conf/icse/HorwitzR92} represent dependencies between elements in programs. We distinguish control and data dependencies. This information is important, for example, to detect whether a loop boundary depends on an input variable (as is the case in program $P_{\mathit{SUM}}$). 
\item[Abstract syntax trees:] ASTs reflect the syntactical structure of programs according to a given grammar and can for instance help to reveal the complexity of expressions. 
\end{description} 

\begin{figure}[t]
\begin{verbatim}
1  int i = 0;               1   int i = 0;  					
2  int n = abs(input());    2   int n = abs(input()); 	             				
3  while (i < n)            3   while (i < n)      				
4         i++;              4          i++;      			
5  assert (i == n);         5   assert (i != n);          
\end{verbatim} 
\caption{Two programs indistinguishable by our kernel}
\label{fig:2programs}
\end{figure}

\noindent Unlike CFGs and PDGs but (partly) alike ASTs, we abstract from concrete names occuring in programs. Nodes in the graph will thus not be labelled with statements or variables as occuring in the program, but with abstract identifiers. We let $Lab$ be the set of all such labels. Table \ref{tab:ids} lists some identifiers and their meaning. The following definition formalizes this graph representation.\footnote{Actually, it only partly does, because a full formalization would require definitions of CFGs, PDGs and ASTs which -- due to lack of space -- cannot all be given here.}

\begin{definition} Let $P$ be a verification task. The {\em graph representation} of $P$ is a graph $G =(N,E,s,t,\rho,\tau,\eta)$ with
\begin{itemize}
\item $N$ a set of nodes (basically, we build an AST for every statement in $P$, and use the nodes of these ASTs),
\item $E$ a set of edges, with $s: E \rightarrow N$ denoting the start and $t: E \rightarrow N$ the end node of an edge,
\item $\rho : N \rightarrow Lab$ a labelling function for nodes, 
\item $\tau : E \rightarrow \{\mathit{CD},\mathit{DD}, \mathit{SD}, \mathit{CF}\}$ a labelling function for edges reflecting the type of dependence: $\mathit{CD}$ (control dependency) and $\mathit{DD}$ (data dependency) origin in PDGs, $\mathit{SD}$ (syntactical dependence) is the ``consists-of'' relationship of ASTs and $\mathit{CF}$ (control flow) the usual control flow in programs, 
\item $\nu: E \rightarrow \{T,F \}$ a function labelling control dependence edges according to the valuation of the conditional they arise from. All other edges are labelled true. 
\end{itemize}
We let $\cG_V$ denote the set of all verification task graphs. 
\end{definition} 

\noindent Figure \ref{fig:graph-sum} depicts the graph representation of the verification task $P_{\mathit{SUM}}$.   The rectangle nodes represent the statements in the program and act as root nodes of small ASTs.  For instance, the rectangle labelled \texttt{Assert} at the bottom, middle represents the assertion in line 10. The gray ovals represent the AST parts below the root nodes.  We define the {\em depth} of nodes $n$, $d(n)$, as the distance of a node to its root node. As an example, the depth of the \texttt{Assert}-node itself is 0, the depth of both {\texttt{==}-nodes is 2.

This graph representation allows us to see the key {\em structural} properties of a verification task, e.g., that the loop (condition) in our example program depends on an assignment where the right-hand-side is an input (which makes verification more complicated). With respect to {\em semantical} properties, our graph representation (as well as all feature-based approaches relying on static analyses of programs) is less adequate. To see this, consider the two programs in Figure \ref{fig:2programs}. They only differ in the assertion at line 5, which from its syntax is the same on both sides: a simple boolean expression on two variables of exactly the same type and dependencies. However, verification of the left program is difficult for verification tools which cannot generate loop invariants. Verification of the program on the right, however, is easy as it is incorrect (which can e.g.\ be detected by a bounded unrolling of the loop). Here, we clearly see the limits of any learning approach based on structural properties of programs.

\section{Predicting Rankings}

This section starts with a short description of the necessary background in machine learning. More specifically, we explain the problem of label ranking as well as the method of ranking by pairwise comparison for solving this problem. In the second part, we recall binary classification with support vector machines and introduce our kernel functions on verification tasks.

\subsection{Label Ranking}

Consider a finite set of $K$ alternatives identified by \emph{class labels} $\cY = \{ y_1, \ldots , y_K\}$; in our case, the alternatives correspond to the verification tools to be compared. We are interested in total order relations $\succ$ on $\cY$, that is, complete, transitive, and antisymmetric relations, where $y_i \succ y_j$ indicates that $y_i$ precedes $y_j$ in the order. Formally, a total order $\succ$ can be identified with a permutation $\pi$ of the set $[K]= \{1 , \ldots , K \}$, such that $\pi(i)$ is the position of $y_i$ in the order.
We denote the class of permutations of $[K]$ (the symmetric group of order $K$) by $\mathbb{S}_K$. By abuse of terminology, though justified in light of the above one-to-one correspondence, we refer to elements $\pi \in \mathbb{S}_K$ as both permutations and rankings.

In the setting of label ranking (see e.g.\ \cite{DBLP:books/daglib/p/VembuG10}), preferences on $\cY$ are ``contextualized'' by instances $\vec{x} \in \cX$, where $\cX$ is an underlying instance space; in our case, instances are programs to be verified. Thus, each instance $\vec{x}$ is associated with a ranking $\succ_{\vec{x}}$ of the label set $\cY$ or, equivalently, a permutation $\pi _{\vec{x}} \in \mathbb{S}_K$. More specifically, since label rankings do not necessarily depend on instances in a deterministic way, each instance $\vec{x}$ is associated with a probability distribution $\mathbf{P}( \cdot \, \vert \,  \vec{x})$ on $\mathbb{S}_K$. Thus, for each $\pi \in \mathbb{S}_K$, $\mathbf{P}( \pi \, \vert \, \vec{x})$ denotes the probability to observe the ranking $\pi$ in the context specified by $\vec{x}$. 

The goal in label ranking is to learn a ``label ranker'', that is, a model 
$$
\mathcal{M}: \, \cX \rightarrow \mathbb{S}_K
$$ 
that predicts a ranking $\hat{\pi}$ for each instance $\vec{x}$ given as an input. More specifically, seeking a model with optimal prediction performance, the goal is to find a risk (expected loss) minimizer 
$$
\mathcal{M}^* \in \argmin_{\mathcal{M} \in \mathbf{M}}
\int_{\cX \times \mathbb{S}_K} D(\mathcal{M}(\vec{x}) ,  \pi) \, d \, \mathbf{P} \enspace ,
$$
where $\mathbf{M}$ is the underlying model class, $\mathbf{P}$ is the joint measure $\mathbf{P}(\vec{x},\pi) = \mathbf{P}(\vec{x}) \mathbf{P}( \pi \given \vec{x})$ on $\cX \times \mathbb{S}_K$ and $D$ is a loss function on $\mathbb{S}_K$. A common example of such a loss is $D(\pi , \hat{\pi} ) = 1 - S(\pi , \hat{\pi} )$, where $S(\pi , \hat{\pi} )$ is the  {\em Spearman rank correlation} \cite{Spear}:
\[ 
S(\pi , \hat{\pi} ) = 
1 - \frac{6\  \sum_{i=1}^K (\pi(i) - \hat{\pi}(i))^2}{K (K^2 -1)} \in [-1,1] 
\]
As training data $\mathbb{D}$, a label ranker uses a set of instances ${\vec{x}}_n$ ($n \in [N]$), together with information about the associated rankings $\pi_n$.

\subsection{Ranking by Pairwise Comparison}

Ranking by pairwise comparison (RPC) is a meta-learning technique that reduces a label ranking task to a set of binary classification problems \cite{rpc08}. More specifically, the idea is to train a separate model (base learner) $\set{M}_{i,j}$ for each pair of labels $(y_i, y_j) \in
\cY$, $1 \leq i < j \leq K$; thus, a total number of $K(K-1)/2$
models is needed (see Figure \ref{fig:rp*c} for an illustration).

For training, the original data $\mathbb{D}$ is first turned into binary classification data sets $\mathbb{D}_{i,j}$, $1 \leq i < j \leq K$. To this end, each preference information of the form
$y_i \succ_x y_j$ (extracted from full or partial information about a ranking $\pi_x$) is turned into a positive (classification)
example $(x,1)$ for the learner $\set{M}_{i,j}$; likewise, each preference $y_j \succ_x y_i$ is turned into a negative 
example $(x,0)$. Thus, $\set{M}_{i,j}$ trained on $\mathbb{D}_{i,j}$ is intended to learn the mapping that outputs 1
if $y_i \succ_x y_j$ and 0 if $y_j \succ_x y_i$.
This mapping  can be realized by any binary classifier.
Instead of a $\{0,1\}$-valued classifier, one can of course
also employ a scoring classifier.
For example, the output of a probabilistic classifier would be a number in
the unit interval $[0,1]$ that can be interpreted as a probability 
of the preference $y_i \succ_x y_j$.
In our approach, we use support vector machines as base learners \cite{vapn_sl98,scho_lw} .

\begin{figure}[t]
\begin{center}
\includegraphics[width=0.45\textwidth]{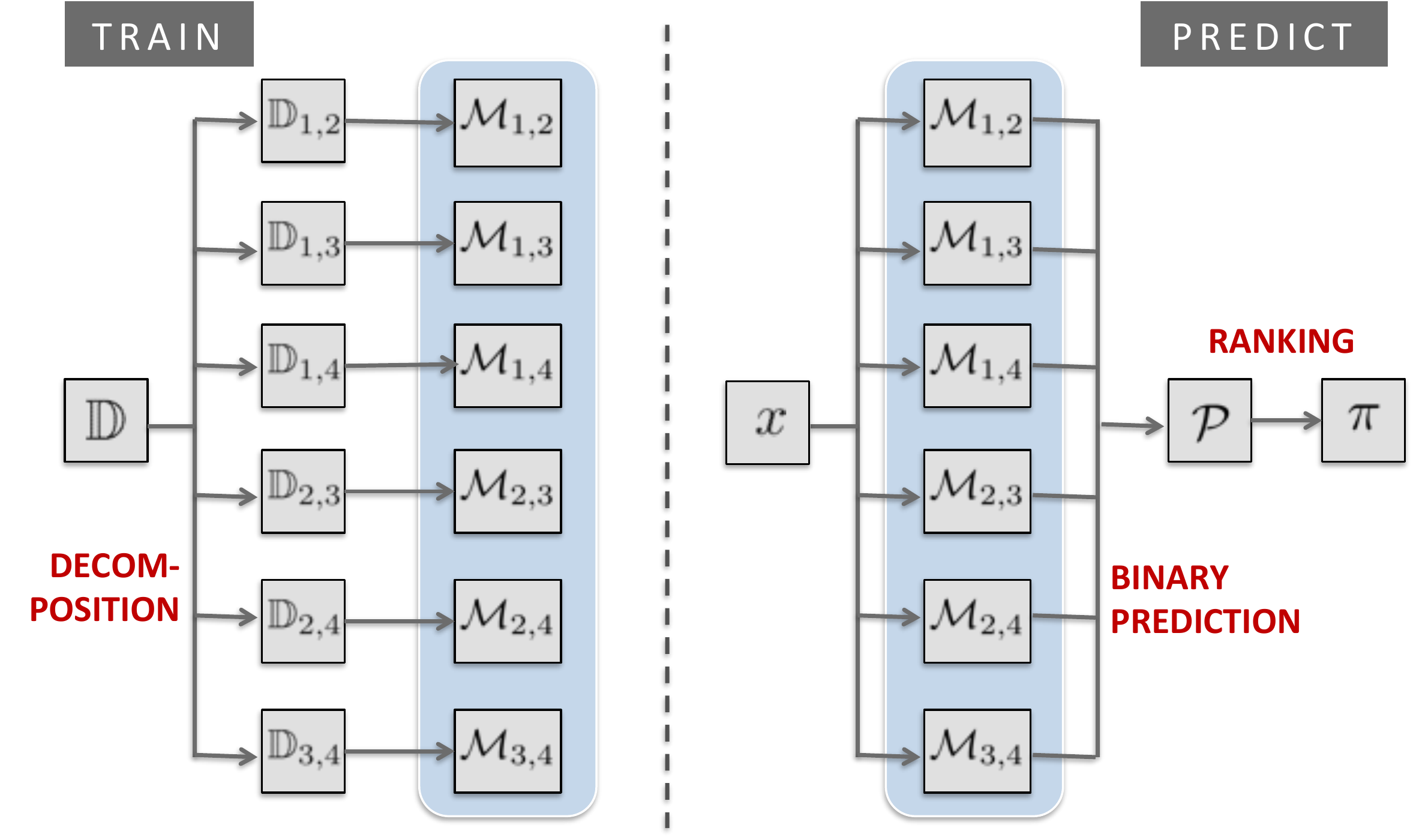}
\end{center}
\caption{Illustration of the RPC approach (for $K=4$). At training time (left), the original data $\mathbb{D}$ is split into $K(K-1)/2$ smaller data sets, one for each pair of labels, and a binary classifier is trained on each of these data sets. If a prediction for a new instance is sought (right), this instance is submitted to each of the binary models, and the pairwise preferences obtained as predictions are combined into a complete ranking $\pi$ via a ranking procedure $\mathcal{P}$. }
\label{fig:rp*c}
\end{figure}

At classification time, a query $x_0 \in \cX$ is submitted to the complete ensemble of
binary learners. Thus, a collection of predicted pairwise preference degrees
$\set{M}_{i,j}(x)$, $1 \leq i, j \leq K$, is obtained. 
The problem, then, is to turn these pairwise preferences into a ranking of the label set $\cY$.
To this end, different ranking procedures can be used. 
The simplest approach is to extend the (weighted)
voting procedure that is often applied in pairwise classification \cite{furn_rr02}: 
For each label $y_i$, a score
\[
S_i \, = \, \sum_{1 \leq j \neq i \leq K}
\set{M}_{i,j}(x_0) 
\]
is derived (where $\set{M}_{i,j}(x_0) = 1- \set{M}_{j,i}(x_0)$ for $i > j$), and then the
labels are sorted according to these scores. Despite its simplicity,
this ranking procedure has several appealing properties. Apart from its
computational efficiency,  it turned out to be relatively robust in practice and, moreover, it possesses some
provable optimality properties in the case where Spearman's rank correlation 
is used as an underlying accuracy measure \cite{jcss10}. 

%Roughly speaking, if the binary learners are unbiased probabilistic classifiers, the simple  ``ranking by weighted voting'' procedure  yields a label ranking that maximizes the expected Spearman rank correlation \citep{mpub168}. Finally, it is worth mentioning that, by changing the ranking procedure, the pairwise approach can also be adjusted to accuracy measures other than Spearman's rank correlation.

\subsection{Support Vector Machines}
 
As already said, support vector machines (SVMs) are used as base learners in RPC. SVMs are so-called ``large margin'' classifiers \cite{scho_lw}. They separate positive from negative training instances in $\mathbb{R}^m$ by means of a linear hyperplane that maximizes the minimum distance of any of the training instances from the hyperplane (decision boundary). Formally, a hyperplane $\{ \vec{x} \with \vec{w}^\top \vec{x} + b = 0 \}$ in $\mathbb{R}^m$ is characterized by the normal vector $\vec{w} \in \mathbb{R}^m$ and the bias term $b \in \mathbb{R}$. Then, encoding the two classes by $\pm 1$, the margin of a training example $(x_i,y_i) \in \mathbb{R}^m \times \{ -1 , + 1 \}$ is given by $y_i ( \vec{w}^\top \vec{x}_i  + b)$; thus, a positive margin indicates that $x_i$ is on the right side of the decision boundary, and hence classified correctly, whereas a negative margin corresponds to a mistake on the training data.

The ``soft margin'' version allows for adding a slack variable $\xi_i \geq 0$ and defines the margin as $y_i ( \vec{w}^\top \vec{x}_i  + b)  + \xi_i$ for each example $x_i$; this is necessary in the case of data that is not linearly separable. Obviously, the values of the slack variables should be kept small, i.e., the problem comes down to finding a reasonable balance between a large (soft) margin and a small amount of slack.  This problem can be formalized in terms of a constrained quadratic optimization problem:
$$
(\vec{w}^*, b^*) = \argmin_{\vec{w}, b, \xi} \left\{  \frac{1}{2} \| \vec{w} \|^2 + C \sum_{i=1}^N \xi_i  \right\} 
$$
subject to the constraints 
\begin{equation}\label{eq:svmcr}
y_i ( \vec{w}^\top \vec{x}_i  + b)  \geq  1- \xi_i \, , \quad \xi_i \geq 0 \, ,
\end{equation}
where $C$ is a parameter that controls the penalization of errors on the training data (indicated by a non-zero $\xi_i$). Instead of solving this problem directly, it is often more convenient to solve its dual.

At prediction time, a new instance $x_0 \in \mathbb{R}^m$ is classified positive or negative depending on whether it lies above or below the hyperplane $(\vec{w}^*, b^*)$. Instead of only returning a binary decision, the distance from the hyperplane is often reported as kind of measure of certainty (with the idea that the closer an instance to the decision boundary, the less certain the prediction). As a disadvantage of this measure, note that the distance is not normalized and therefore difficult to interpret and compare. So-called Platt scaling is a post-processing step, in which distances are mapped to $[0,1]$ via a logistic transformation; thus, each instance is assigned a (pseudo-)probability of belonging to the positive class \cite{platt}.

In the dual formulation of the above optimization problem, training instances $x_i, x_j$ never occur in isolation but always in the form of inner products $\langle x_i , x_j \rangle$. This allows for the ``kernelization'' of SVMs, simply be replacing such inner products by values $k(x_i, x_j)$ of a so-called kernel function $k(\cdot)$. 
\begin{definition} A function $k: \cX \times \cX \rightarrow \mathbb{R}$ is a {\em positive semi-definite kernel} iff $k$ is symmetric, i.e., $k(x,x') = k(x',x)$,  and 
\[ 
\sum_{i=1}^N \sum_{j=1}^N c_i c_j k(x_i,x_j) \geq 0 \]
for arbitrary $N$, arbitrary instances $x_1, \ldots, x_N \in \cX$ and arbitrary $c_1, \ldots, c_N \in \mathbb{R}$.
\end{definition}
If $k(\cdot)$ is a proper kernel function, one can guarantee the existence of an induced feature space $\mathcal{F}$ (which is a Hilbert space) and a feature map $\phi: \, \mathcal{X} \rightarrow \mathcal{F}$ such that $\langle \phi(x) , \phi(x') \rangle = k(x_i, x_j)$. Thus, the computation of inner products in the (typically very high-dimensional) space $\mathcal{F}$ can be replaced by the evaluations of the kernel, which in turn allows a linear model to be fit in $\mathcal{F}$ without ever accessing that space or computing the image $\phi(x_i)$ of a training instance $x_i$---this is called the ``kernel trick''. The learning algorithm only requires access to the \emph{Gram matrix}, i.e., the value of the kernel for each pair of training instances:
$$
G = \left( 
\begin{array}{cccc}
k(x_1, x_1) & k(x_1, x_2) & \ldots & k(x_1, x_N) \\
k(x_2, x_1) & k(x_2, x_2) & \ldots & k(x_2, x_N) \\
\vdots & \vdots & \ddots & \vdots \\
k(x_N, x_1) & k(x_N, x_2) & \ldots & k(x_N, x_N) 
\end{array}
\right)
$$
Note that the instance space $\cX$, on which the kernel is defined, is not necessarily an Euclidean space. Instead, $\cX$ can be any space or set of objects. In particular, this allows SVMs to be trained on \emph{structured} (non-vectorial) objects. In general, a kernel function can be interpreted as a kind of similarity measure on $\cX$, i.e., the more similar instances $x_i, x_j$, the larger $k(x_i,x_j)$. Next, we address the question of how to define appropriate kernel functions on verification tasks.

\subsection{Graph Kernels for Verification Tasks} 

Verification tasks are represented by specific graphs, whence our kernel needs to operate on graphs. A number of graph kernels already exist, for instance based on comparisons of shortest paths or random walks of graphs. However, most of these graph kernels do not scale well to large graphs \cite{DBLP:journals/jmlr/ShervashidzeSLMB11}. As our graphs are representations of programs with several  thousands lines of code, and hence very large, we have chosen to proceed from our own kernel development based on {\em Weisfeiler-Lehman subtree kernels} \cite{DBLP:journals/jmlr/ShervashidzeSLMB11}, which are known to scale better. 

\begin{algorithm}[t]
\caption{relabel (Graph relabelling)}\label{alg:relabel}
\begin{algorithmic}[1]
\Require 
\Statex $G = (N,E,s,t,\rho,\tau,\nu)$  graph 
\Statex $z: \Sigma^* \rightarrow \Sigma$ injective compression function 
\Statex $\eta: N\rightarrow 2^E$ neighbour function 
\Statex $m$ iteration bound
\Ensure 
\Statex relabelled  graph $G$
%\State $G^0 := G$ 
%\State $\forall n \in N: str(n):= \rho(n)$ 
\For {$i = 1$ to $m$}
\For{ $n \in N$} 
\State $Aug(n):=\big \langle z \big(\rho(s(e)) \oplus \tau(e) \oplus \nu(e)\big) \mid e \in \eta(n) \big \rangle $
\State $Aug(n):=sort(Aug(n))$
\State $str(n):=concat(Aug(n))$ 
\State $str(n):=\rho(n)\oplus str(n)$
\State $\rho(n):=z(str(n))$ 
\EndFor
\EndFor 
\State \Return $G$
\end{algorithmic}
\end{algorithm} 

%\fbox{brauchen wir jedes mal einen neuen Graph oder reicht einer?}
Weisfeiler-Lehman kernels are extensions of the Weisfeiler-Lehman test of isomorphism between two discretely labelled, undirected graphs \cite{WeisLeh68}. This test basically compares graphs according to their node labels.  For taking edges into account, node labels are extended with information about neighbouring nodes in three steps:
\begin{description}
\item[Augmentation:] Concatenate label of node $n$ with labels of its neighbours,
\item[Sorting:] Sort this sequence according to predefined order on labels,
\item[Compression:] Compress thus obtained sequences into new labels. 
\end{description} 
\noindent These steps are repeated until the node label sets of the two graphs differ or until a predefined bound on the number of iterations is exhausted. This bound is used to regulate the depth of subtrees considered. Note that this is a test only, not a proof of isomorphism.  

For making this Weisfeiler-Lehman test act as a kernel for verification tasks, we made three adaptations to the graph relabelling, giving  rise to Algorithm \ref{alg:relabel}: 
\begin{itemize}
\item[(1)] extension to directed multigraphs, 
\item[(2)] customization to specific neighbours of nodes, and 
\item[(3)] integration of edge labels. 
\end{itemize}
In Algorithm \ref{alg:relabel}, we use the notation $\langle \ldots \mid \ldots \rangle$ for list comprehensions, defining a sequence of values. Moreover, $z$ is the {\em compression function} compressing sequences of labels into new labels (which thus needs to be injective). In our case, we use numbers as labels, i.e., $\Sigma=\mathbb{N}$ with the usual ordering $\leq$. To this end, we first map all node identifiers and edge labels to $\mathbb{N}$. Every newly arising sequence then simply gets a new number assigned. 
The {\em neighbour function} $\eta$ is used to customize kernels by selectively choosing the neighbours to be considered during augmentation. Thereby, we can specialize our kernels to just control flow or just data dependence edges, for example. The functions $sort$ and $concat$ sort sequences of labels (in ascending order) and concatenate sequences, respectively. 
%We let $G^m$ be the graph obtained from the algorithm when given $G$ as input graph and $m$ as iteration depth. \fbox{die anderen Parameter? compress? sort?} and fixing some $\eta$?

This lets us finally define our kernels for verification tasks. 

\begin{definition} 
Let $G_i = (N_i, E_i, s_i, t_i, \rho_i, \tau_i, \nu_i)$, $i=1,2$ be graph representations of verification tasks, $z : \Sigma^* \rightarrow \Sigma$ a compression function, $m \in \mathbb{N}$ an iteration bound, $d\in \mathbb{N}$ a depth for subtrees and $\eta_i: N_i \rightarrow 2^{E_i}$ neighbour functions. The {\em verification graph kernel } $k_{\eta_1,\eta_2,z}^{(d,m)} : \cG_V \times \cG_V \rightarrow \mathbb{R}$ is defined as 
\[
  k_{\eta_1,\eta_2,z}^{(d,m)}(G_1,G_2)  =  \sum_{i=1}^m k^d\left(\begin{array}{l}relabel(G_1,z,\eta_1,m),\\ relabel(G_2,z,\eta_2,m)\end{array} \right)
\]
\mbox{ with }
\begin{eqnarray*}
  k^d(G,G') & = & \sum_{n\in N} \sum_{n'\in N'} k^d_\delta(n,n') \mbox{ and } \\
  k^d_\delta(n,n') & = & \left\{ \begin{array}{ll} \delta(\rho_1(n),\rho_2(n')) & \mbox{ if } d(n) \leq d \wedge d(n') \leq d \\
  0 & \mbox{ else }
  \end{array},\right.
\end{eqnarray*}

\noindent where $\delta$ is a Dirac kernel defined as $\delta(u,w) = 1$ if $u$ equals $w$ and 0 otherwise. 
\end{definition} 

\noindent Intuitively, the kernels count the number of equally labelled nodes in all iterations, where the iteration bound steers to what extent subtrees of root ASTs nodes are considered, the neighbour function controls what edges are taken into account, and the depth $d$ fixes whether a node is considered at all. For the latter, remember that the depth of a node is its distance to its top-level AST node. By incorporating the depth, we have the option to consider or ignore details of expressions. 

\noindent We can show the following result (for the proof, we refer to \cite{MC16}):

\begin{theorem}
The kernel $k_{\eta_1,\eta_2,z}^{(d,m)}$  is positive  semi-definite.
\end{theorem}

Our kernels can now be used in a support vector machine within the ranking by pairwise composition approach outlined above.

\section{Implementation and Experimental Evaluation}

\begin{table*}[t]
    \centering
        \caption{SV-COMP 2015 -- Prediction accuracy (mean and standard deviation) in terms of the Spearman rank correlation (higher is better, minimum is $-1$, maximum is $+1$).}
    \label{tab:accuracy}
    \begin{tabular}{|r||c|c|c|}
        \hline
        Kernel / Data Set & SAFETY & TERMINATION & MEMSAFETY \\ \hline \hline
        $k_{\mathit{CF}}$ (CFG) & $\bm{.616 \pm .002}$ & $.709 \pm .008$ & $.774 \pm .003$ \\ \hline
        $k_{\mathit{DD}}$ (data dependency) & $.615 \pm .004$ & $.680 \pm .009$ & $.761 \pm .003$ \\ \hline
        $k_{\mathit{CD}}$ (control dependency) & $.607 \pm .005$ & $.674 \pm .008$ & $.767 \pm .005$ \\ \hline
        $k_{\mathit{CD,DD}}$ (PDG) & $.611 \pm .005$ & $.692 \pm .006$ & $.771 \pm .009$ \\ \hline
        $k_{\mathit{CF,CD,DD}}$ (PDG + CFG) & $.614 \pm .004$  & $.692 \pm .002$  & \bm{$.784 \pm .003$}  \\ \hline
        weighted combination & $.615 \pm .002$ & $\bm{.711 \pm .007}$ & $.769 \pm .004$ \\ \hline 
        features of \cite{DBLP:conf/cav/DemyanovaPVZ15} & $.576 \pm .006$ & $.609 \pm .011$ & $.692 \pm .006$ \\ \hline
				default predictor & $.560 \pm .003$ & $.332 \pm .008$ & $.604 \pm .002$ \\ \hline
    \end{tabular}
\end{table*}

\begin{table*}[t]
    \centering
        \caption{SV-COMP 2017 -- Prediction accuracy (mean and standard deviation) in terms of the Spearman rank correlation (higher is better, minimum is $-1$, maximum is $+1$).}
    \label{tab:accuracySVCOMP2017}
    \begin{tabular}{|r||c|c|c|}
        \hline
        Kernel / Data Set & SAFETY & TERMINATION & MEMSAFETY \\ \hline \hline
        $k_{\mathit{CF}}$ (CFG) & $\bm{.635 \pm .003}$ & $.657 \pm .007$ & $.755 \pm .004$ \\ \hline
        $k_{\mathit{DD}}$ (data dependency) & $ .618\pm .003$ & $.635 \pm .006$ & $.754 \pm .007$ \\ \hline
        $k_{\mathit{CD}}$ (control dependency) & $ .627\pm.002 $ & $.637 \pm .006$ & $.756 \pm .005$ \\ \hline
        $k_{\mathit{CD,DD}}$ (PDG) & $ .630\pm $ .005& $.644 \pm .003$ & \bm{$.757 \pm .007$} \\ \hline
        $k_{\mathit{CF,CD,DD}}$ (PDG + CFG) & $ .632 \pm.004$  & $.658 \pm .009$  & $.756 \pm .003$  \\ \hline
        weighted combination & $ .634 \pm .003$ & $\bm{.664 \pm .010}$ & $.756 \pm .003$ \\ \hline 
        features of \cite{DBLP:conf/cav/DemyanovaPVZ15} & $ .560 \pm .004 $ & $ .560 \pm .006 $ & $.717 \pm .001$ \\ \hline
				default predictor & $ .452 \pm .003 $ & $ .339 \pm .004 $ & $.668 \pm .001$ \\ \hline
    \end{tabular}
\end{table*}

In our experiments, we studied the performance of our method for rank prediction
%More specifically, we determined whether predicted rankings reflect the observed rankings 
in the SV-COMP 2015 and 2017. 
% Furthermore, we examined the time for rank prediction, since predictions must be fast to be useful in practice.
To this end, we compared six variants of our kernel with respect to prediction accuracy, each of which focuses on different aspects of a program. Such kind of customization of kernels becomes possible  thanks to the two neighbouring functions $\eta_1$ and $\eta_2$. In our case, neighbours are chosen according to the type of edge connecting them. We define $\eta_\ell, \ell \in \{\mathit{CD},\mathit{DD}, \mathit{SD}, \mathit{CF}\} $ to be $\eta_\ell(n) = \{ e \mid \tau(e) = \ell \wedge s(e) = n\}$, and let $\eta_L(n) = \bigcup_{\ell \in L} \eta_\ell(n)$ for a node $n$. For our kernels, we always use the same neighbouring function on both graphs. Hence, we will just use the edge labels employed in neighbouring functions as indizes for kernels. 

Our experiments include kernels 
\begin{itemize}
\item $k^{(d,m)}_{\{\mathit{CF}\}}$ (control-flow), 
\item $k^{(d,m)}_{\{CD\}}$ (control dependencies), 
\item $k^{(d,m)}_{\{\mathit{DD}\}}$ (data dependencies), 
\item $k^{(d,m)}_{\{\mathit{CD, DD}\}}$ (control and data dependencies), and 
\item $k^{(d,m)}_{\{\mathit{CF, CD, DD}\}}$ (control-flow, data- and control-\\dependencies). 
\end{itemize}

\noindent In addition, we included an equally weighted linear combination $k^{(d,m)}_{lin}$ of some of our kernels, which is defined as 
\[\begin{array}{l}
   k^{(d,m)}_{lin}(G_1, G_2) =\\
	\frac{1}{3} k^{(d,m)}_{\{\mathit{CF}\}}(G_1, G_2) + \frac{1}{3} k^{(d,m)}_{\{\mathit{CD}\}}(G_1, G_2) + \frac{1}{3} k^{(d,m)}_{\{\mathit{DD}\}}(G_1, G_2) 
\end{array}\]
(one can easily check that this is again a valid kernel, see e.g.\ \cite{MC16}).
To get an insight on how the prediction accuracy performs compared to state-of-the-art approaches, we also included the accuracy achieved by using the feature vectors from Demyanova et al.~\cite{DBLP:conf/cav/DemyanovaPVZ15}. In addition, we constructed a {\em default predictor} for comparison: the default predictor takes all rankings of the data set used for learning, determines the ranking which minimizes the distance (wrt.\ Spearman rank correlation) to these rankings and always predicts this default ranking without any learning.

We constructed the following data sets for our experiments: SAFETY, TERMINATION, and MEMSAFETY.
Each data set consists of several verification tasks taken from the SV-COMP 2015 and 2017 benchmark sets.
To provide a comprehensive analysis under varying conditions, each data set represents a different property type (safety, termination, and memory safety).
In case of SV-COMP 2015, SAFETY is a data set of 483 verification tasks originating from the SV-COMP categories \textit{Loops}, \textit{BitVectors}, \textit{Floats}, \textit{Simple}, \textit{ControlFlowInteger}, and \textit{HeapManipulation}. For 2017, our SAFETY set consists of 637 verification tasks out of the categories \textit{ReachSafety-Bitvectors}, \textit{ReachSafety-ControlFlow}, \textit{ReachSafety-Heap}, \textit{ReachSafety-Floats} and \textit{ReachSafety-Loops}.  
The set of considered tools in SAFETY consists of the tools, which participated in all these categories (6 tools for 2015 and 11 tools for 2017).
TERMINATION is a data set of verification tasks taken from the category \textit{Termination}, 393 tasks in 2015 and 507 for 2017. 
In this data set, we consider tools that participated in this category and successfully proved or disproved termination on at least one verification task (5 tools both for 2015 and 2017).
MEMSAFETY is the data set of verification tasks consisting of tasks from the category \textit{MemorySafety}, 205 for 2015 and 181 for 2017. 
Again, we considered only tools that report at least one correct outcome (9 tools in 2015 and 11 tools in 2017).

For the computation of our verification graphs, we used the configurable software analysis framework \caps{CPAChecker} \cite{DBLP:conf/cav/BeyerK11}:
To obtain control-flow and AST information, we used the integrated C parser.
In case of data dependencies, we utilized the integrated reaching definition analysis as is described in \cite{DBLP:conf/icse/HorwitzR92}.
For the sake of simplicity, we ignored complex dependencies introduced by pointers.
Also according to \cite{DBLP:conf/icse/HorwitzR92}, we computed control dependencies. 
Eventually, we built another extension of the \caps{CPAChecker} that combines all the collected information into one graph using the JGraphT library\footnote{\url{http://jgrapht.org}}.
To solve our label ranking problem, we integrated the RPC approach and our kernel framework into the scikit-learn library\footnote{\url{http://scikit-learn.org}}. 
There, we also employed the implementation of support vector machines (with Platt scaling) offered by scikit-learn.
Finally, we integrated the feature vectors of \cite{DBLP:conf/cav/DemyanovaPVZ15} through the tool Verifolio\footnote{\url{http://forsyte.at/software/verifolio/}}. 
All the code and data (of 2015) is available via GitHub\footnote{\url{https://github.com/zenscr/PyPRSVT}}.

\begin{table*}[t]
    \centering
        \caption{SV-COMP 2015 -- Time (in seconds) for training and testing (mean $\pm$ standard deviation)}
    \label{tab:time}
    \begin{tabular}{|r|r||c|c|c|}
        \cline{2-5} 
        \multicolumn{1}{r|}{}  & Kernel / Data Set & SAFETY & TERMINATION & MEMSAFETY \\  \hline 
        \multirow{7}{*}{\begin{sideways} training~ \end{sideways}} & $k_{\mathit{CF}}$ (CFG) & $367 \pm .634$ & $147 \pm 1.10$ & $128 \pm .728$ \\ \cline{2-5} 
        & $k_{\mathit{DD}}$ (data dependency) & $461 \pm 13.8$ & $172 \pm 4.74$ & $137 \pm 1.81$ \\ \cline{2-5}
        & $k_{\mathit{CD}}$ (control dependency) & $393 \pm 2.14$ & $160 \pm 2.32$ & $129 \pm 6.35$ \\ \cline{2-5}
        & $k_{\mathit{CD,DD}}$ (PDG) & $359 \pm 1.53$ & $131 \pm .627$ & $126 \pm .445$ \\ \cline{2-5}
        & $k_{\mathit{CF,CD,DD}}$ (PDG + CFG) & $318 \pm 5.06$  & \bm{$125 \pm 1.54$}  & $125  \pm 2.72$  \\ \cline{2-5}
        & weighted combination & $349 \pm 9.13$ & $127 \pm 1.17$ & \bm{$120 \pm .200$} \\ \cline{2-5}
        & features of \cite{DBLP:conf/cav/DemyanovaPVZ15} & \bm{$266 \pm 3.55$} & $138 \pm .282$ & $150 \pm .410$ \\ \hline
        \hline
        \multirow{7}{*}{\begin{sideways} testing~ \end{sideways}} & $k_{\mathit{CF}}$ (CFG) & $.021 \pm .0$ & $.011 \pm .0$ & $.035 \pm .0$ \\ \cline{2-5} 
        & $k_{\mathit{DD}}$ (data dependency) & $.020 \pm .0$ & $.011 \pm .0$ & $.034 \pm .0$ \\ \cline{2-5}
        & $k_{\mathit{CD}}$ (control dependency) & $.021 \pm .0$ & $.011 \pm .0$ & $.034 \pm .0$ \\ \cline{2-5}
        & $k_{\mathit{CD,DD}}$ (PDG) & $.020 \pm .0$ & $.012 \pm .0$ & $.034 \pm .0$ \\ \cline{2-5}
        & $k_{\mathit{CF,CD,DD}}$ (PDG + CFG) & $.022 \pm .005$  & $.013 \pm .0$  & $.036  \pm .0$  \\ \cline{2-5}
        & weighted combination  & $.019 \pm .0$ & $.011 \pm .0$ & $.033 \pm .0$ \\ \cline{2-5}
        & features of \cite{DBLP:conf/cav/DemyanovaPVZ15} & $.026  \pm .0$ & $.013 \pm .0$ & $.019 \pm .0$ \\ \hline
        \hline
    \end{tabular}
\end{table*}

To examine the prediction accuracy for each configuration, we performed a 10-fold cross-validation. A $k$-fold \textit{cross-validation} is a commonly used technique for model assessment. First, the data is divided into $k$ subsets of equal size. Then, one subset is used as test set, whereas the learning algorithm trains a model on the remaining $k-1$ subsets. This procedure is repeated exactly $k$ times, each time using one of the folds for testing, and the overall performance is obtained as the average of the $k$ test performances thus produced.  After each step of the cross-validation, we compared the actual true rankings on the test sets to the corresponding predicted rankings (with RPC and SVMs) using the Spearman rank correlation.
The overall \textit{accuracy} is then the average over all the accuracies encountered in each step.

%The input of each SVM always consisted of a gram matrix computed with one of our kernels and the corresponding binary labels as given by the RPC approach. In addition, all the hyperparameters an SVM depends upon (e.g.\ our kernel parameters $d$ and $m$) were optimized using grid-search again via multiple cross-validation runs on the training data.
%Note that this nested cross-validation prevented an optimistic bias in the model evaluation, because the selection of hyperparameters is always performed on the training set only.
%Each SVM is eventually trained on the full training set (of the outer cross-validation) using the selected hyperparameters.
%In order to exclude random effects caused by the separation of the training data into folds, we repeated the whole experiment five times (i.e., we conducted five 10-fold cross validations, each time with different separations into folds).
%Further, we always precomputed the matrices before performing cross-validation. 

In Table \ref{tab:accuracy} (SV-COMP 2015) and Table \ref{tab:accuracySVCOMP2017} (SV-COMP 2017), we report the average prediction accuracies (and standard deviations) in terms of the Spearman rank correlation; note that an average accuracy of 0 would be obtained by guessing rankings at random, while $+1$ stands for predictions that perfectly coincide with the true ranking (and $-1$ for completely reversing that ranking). As can be seen, our approach shows a rather strong predictive performance.   Depending on the verification task, different kernels achieve the best results, though the differences in performance are statistically non-significant. More importantly, our approach significantly outperforms the one of \cite{DBLP:conf/cav/DemyanovaPVZ15} as well as the default predictor on all tasks. This applies to the data of 2015 as well as 2017.

Table \ref{tab:time} (SV-COMP 2015) and Table \ref{tab:timeSVCOMP2017} (SV-COMP 2017) show the average training and testing times during the 10-fold cross validation (using the precomputed Gram matrix), i.e., the time in seconds for the training with 9 folds of the input data and the time for computing the rankings (testing) for the remaining fold. As expected, training a model is more time-consuming than using it for prediction. Moreover, like for accuracy, there are no significant differences between the kernels. Interestingly, the kernels are sometimes even faster than the simple feature representation of \cite{DBLP:conf/cav/DemyanovaPVZ15}.

\begin{table*}[t]
    \centering
        \caption{SV-COMP 2017 -- Time (in seconds) for training and testing (mean $\pm$ standard deviation)}
    \label{tab:timeSVCOMP2017}
    \begin{tabular}{|r|r||c|c|c|}
        \cline{2-5} 
        \multicolumn{1}{r|}{}  & Kernel / Data Set & SAFETY & TERMINATION & MEMSAFETY \\  \hline 
        \multirow{7}{*}{\begin{sideways} training~ \end{sideways}} & $k_{\mathit{CF}}$ (CFG) & $ 5584\pm 762$ & $470 \pm 3.92$ & $237 \pm 1.68$ \\ \cline{2-5} 
        & $k_{\mathit{DD}}$ (data dependency) & $ 7229\pm 731$ & $529 \pm 8.11$ & $260 \pm 1.23$ \\ \cline{2-5}
        & $k_{\mathit{CD}}$ (control dependency) & $ 4301\pm 207$ & $510 \pm 7.48$ & $243 \pm 15.2$ \\ \cline{2-5}
        & $k_{\mathit{CD,DD}}$ (PDG) & $  5533\pm490 $ & $440 \pm 10.1$ & $231 \pm 2.0$ \\ \cline{2-5}
        & $k_{\mathit{CF,CD,DD}}$ (PDG + CFG) & \bm{$ 3774\pm 276$}  & \bm{$394 \pm 10.4$}  & $231  \pm 1.23$  \\ \cline{2-5}
        & weighted combination & $ 3809\pm 99.4$ & $413 \pm 5.40$ & \bm{$235 \pm 1.29$} \\ \cline{2-5}
        & features of \cite{DBLP:conf/cav/DemyanovaPVZ15} & $ 4914\pm 273$ & $ 711\pm 21.0$ & $331 \pm 5.25$ \\ \hline
        \hline
        \multirow{7}{*}{\begin{sideways} testing~ \end{sideways}} & $k_{\mathit{CF}}$ (CFG) & $ .347\pm .048$ & $.044 \pm .002$ & $.104 \pm .0$ \\ \cline{2-5} 
        & $k_{\mathit{DD}}$ (data dependency) & $ .341 \pm .020$ & $.043 \pm .002$ & $.106 \pm .003$ \\ \cline{2-5}
        & $k_{\mathit{CD}}$ (control dependency) & $ .290\pm .018$ & $\bm{.038 \pm .002}$ & $.116 \pm .02$ \\ \cline{2-5}
        & $k_{\mathit{CD,DD}}$ (PDG) & $ .424\pm.018 $ & $.039 \pm .002$ & $.104 \pm .001$ \\ \cline{2-5}
        & $k_{\mathit{CF,CD,DD}}$ (PDG + CFG) & \bm{$ .286\pm .007$}  & $.046 \pm .003$  & $.104  \pm .008$  \\ \cline{2-5}
        & weighted combination  & $ .294\pm .022$ & $.048 \pm .005$ & $.112 \pm .011$ \\ \cline{2-5}
        & features of \cite{DBLP:conf/cav/DemyanovaPVZ15} & $  .534\pm .026$ & $ .077\pm .003$ & $\bm{.042 \pm .0}$ \\ \hline
        \hline
    \end{tabular}
\end{table*}

\section{Conclusion}

In the recent years, machine learning has attracted increasing attention in software engineering and related fields, where it has been used, for example, in program construction and analysis.
In this paper, we have proposed a method for predicting rankings of verification tools on given programs. Our rank prediction technique builds on existing methods for label ranking via pairwise comparison. To this end, we have developed an expressive representation of source code, capturing various forms of dependencies between program elements. Instead of explicitly extracting features of programs tailored towards verification, we have constructed a kernel that compares programs according to their elements and the connections between them. Due to its generic nature, we speculate that this kernel will also be useful for other sorts of learning problems on programs---a conjecture we shall verify in future work. 

Our approach can be seen as a tool for \emph{algorithm selection}, a problem that has also been tackled by other authors  \cite{DBLP:conf/cav/DemyanovaPVZ15,DBLP:conf/msr/TulsianKKLN14,DBLP:journals/corr/abs-1111-2249}.
Other applications of machine learning include the learning of programs from examples (\cite{DBLP:conf/popl/RaychevBVK16,Lau}) and the prediction of properties of  programs (e.g., types for program variables \cite{DBLP:conf/popl/RaychevVK15} or malware in Android apps \cite{DBLP:conf/eisic/SahsK12}). Just like our approach, the latter also uses Weisfeiler-Lehman subtree kernels (on CFGs only). A machine learning approach to software verification itself has recently been proposed in \cite{DBLP:conf/icse/ChenHLLTWW16}. However, to the best of our knowledge, the use of machine learning for predicting {\em rankings} of tools (algorithms) has never been tried so far. 
%We can, of course, only predict rankings when we have sufficiently large data sets of tools for training at hand. 

For future work, we are planning to generalize our methodology by exploiting properties (features) of verification tools, which are only identified by their name so far. Recently, a generalization of label ranking called \emph{dyad ranking} has been proposed, in which not only the instances but also the alternatives to be ranked can be described in terms of properties \cite{dyad}. As an important advantage of this approach, note that it in principle allows for ranking alternatives with very few or even no training information so far. This becomes possible by generalizing via the feature descriptions (alternatives with similar properties are expected to perform similarly and hence to occupy similar ranks). In our case, this would mean, for example, that predictions can be made for a new verification tool that has never been tried so far---provided, of course, meaningful descriptions of such tools are available. Developing corresponding representations is one of the challenges we will address next.

 %(algorithm selection for SAT solving) are instances. 

\bibliographystyle{ACM-Reference-Format}
\bibliography{references} 

\end{document}